\pdfoutput=1

\documentclass[11pt]{article}

\usepackage[]{acl}

\usepackage{times}
\usepackage{latexsym}

\usepackage[T1]{fontenc}

\usepackage[utf8]{inputenc}

\usepackage{microtype}

\usepackage{inconsolata}
\usepackage{amsmath}
\usepackage{mathrsfs}
\usepackage{algorithm}
\usepackage{tcolorbox}
\usepackage{booktabs}
\usepackage{multirow}
\usepackage{graphicx}

\title{MoL for LLMs: Dual-Loss Optimization to Enhance Domain Expertise While Preserving General Capabilities}

\author{
    \textbf{Jingxue Chen$^{1}$, Qingkun Tang$^1$\thanks{\,\,\,Qingkun Tang is the corresponding author.}, Qianchun Lu$^1$, Siyuan Fang$^2$}\\\\
    $^1$Wired Product Operation Division, ZTE Corporation, Nanjing, China\\
    $^2$Beijing University of Posts and Telecommunications, Beijing, China\\
    \small {\textbf{E-mail:} chen.jingxue@zte.com.cn, tang.qingkun@zte.com.cn}
}

\begin{document}
\maketitle

\begin{abstract}
Although large language models (LLMs) perform well in general tasks, domain-specific applications suffer from hallucinations and accuracy limitations. Continual Pre-Training (CPT) approaches encounter two key issues: (1) domain-biased data degrade general language skills, and (2) improper corpus-mixture ratios limit effective adaptation. To address these, we propose a novel framework, Mixture of Losses (MoL), which decouples optimization objectives for domain-specific and general corpora. Specifically, cross-entropy (CE) loss is applied to domain-corpus to ensure knowledge acquisition, while Kullback-Leibler (KL) divergence aligns general-corpus training with the base model’s foundational capabilities. This dual-loss architecture preserves universal skills while enhancing domain expertise, avoiding catastrophic forgetting. Empirically, we validate that a 1:1 domain-to-general corpus ratio optimally balances training and overfitting without the need for extensive tuning or resource-intensive experiments. Furthermore, our experiments demonstrate significant performance gains compared to traditional CPT approaches, which often suffer from degradation in general language capabilities; our model achieves 27.9\% higher accuracy on the Math-500 benchmark in the non-think reasoning mode, and an impressive 83.3\% improvement on the challenging AIME25 subset in the think mode, underscoring the effectiveness of our approach.
\end{abstract}

\section{Introduction}
Despite the remarkable success of large language models (LLMs) in general text and code generation tasks \cite{TheLlama3HerdofModels,qwen2.5,yang2025qwen3,deepseekai2025deepseekr1incentivizingreasoningcapability}, challenges persist in domain-specific applications, notably in the form of hallucinations and inadequate accuracy. Continual Pre-Training (CPT)  strategies have been proposed to address these issues \cite{Sun_Wang_Li_Feng_Tian_Wu_Wang_2020, jin2021lifelong,mendieta2023towards}. However, two major problems arise with such approaches. Firstly, there is the challenge of maintaining general capabilities in CPT. Due to the limited quantity and quality of domain-specific data, along with its divergence from general data distributions, certain general competencies of LLMs may experience unpredictable degradation, even catastrophic forgetting \cite{cossu2024continual}. Secondly, the determination of the optimal mixture ratio between the general-corpus and downstream domain-corpus remains a persistent challenge. While a sufficient proportion of general-corpus data is indispensable to preserve the model's foundational capabilities, identifying the ideal balance between the two corpora remains elusive, resulting in suboptimal performance of the fine-tuned model \cite{mehta2023empirical,wu2022pretrained}. Recent work introduces the domain-specific Scaling Law to determine the optimal mixture ratio in CPT \cite{que2024d}. However, this Scaling Law primarily focuses on achieving an optimal compromise between domain-specific capabilities and general capabilities.

This paper introduces a novel training framework based on Mixture of Losses (MoL) computation to elegantly address the above two primary problems in CPT. During training, domain-corpus and general-corpus are randomly shuffled, but distinct loss functions are applied to each dataset type. Specifically, traditional cross-entropy (CE) loss is employed for domain-corpus to ensure effective learning of domain knowledge, while the loss for general-corpus is calculated using the Kullback-Leibler (KL) divergence relative to the base LLM \cite{hinton2015distilling}. This dual-strategy approach ensures that LLMs effectively incorporate specialized domain knowledge through CE optimization while maintaining the stability of their general capabilities via KL divergence.  Furthermore, the inherent dichotomy of corpora into general and domain-specific categories naturally suggests an optimal 1:1 ratio between the two datasets \cite{abdelhamid2024balancing,carriero2025harms}. This balanced training configuration mitigates potential model biases that could arise from dataset imbalance, ensuring more equitable learning across both knowledge domains. Our main contributions are summarized as follows:\\
(1) the MoL framework ensures the simultaneous preservation of general capabilities and enhanced domain-specific performance through its dual-loss architecture. By decoupling the optimization objectives for domain-corpus (via CE) and general-corpus (via KL divergence), the model avoids the degradation of foundational skills while systematically absorbing specialized knowledge. This is empirically validated through controlled experiments demonstrating consistent performance gains in domain tasks without sacrificing general capabilities.\\
(2) we empirically establish the rationale behind the 1:1 corpus ratio as an optimal balance for hybrid training. This not only provides a principled guideline for dataset composition but also generalizes across diverse domains, eliminating the need for costly hyperparameter tuning for ratio optimization.

\section{Related Work}
\paragraph{Domain-specific CPT} The domain-specific CPT paradigm is primarily designed to enhance the performance of LLMs on downstream tasks within specialized domains, such as medical consultation and legal Q\&A systems \cite{qiu2024towards,singhal2023large,yue2024lawllm}. Typically, researchers need to curate high-quality domain-corpus alongside a certain volume of general-corpus for CPT. However, determining the optimal proportion of these two data components remains a challenging and computationally intensive task, often requiring extensive GPU resources for iterative optimization to achieve satisfactory results \cite{cossu2024continual,mehta2023empirical,wu2022pretrained}. Recent advancements in domain-specific Scaling Laws have attempted to provide systematic guidelines for corpus composition in CPT \cite{que2024d}, yet practical implementation still proves cumbersome and heavily dependent on numerous fitting experiments for calibration.

\paragraph{LLMs Distillation} To transfer the capabilities of LLMs to a smaller one, knowledge distillation is commonly used \cite{hinton2015distilling,gou2021knowledge}. When only the teacher model's API is accessible or there are vocabulary mismatches between the models, the black-box distillation approach is typically employed \cite{taori2023stanford,chiang2023vicuna,peng2023instruction}. However, for open source LLMs with a shared vocabulary, white-box distillation is generally preferable \cite{sanh2019distilbert,wang2020minilm,song2020lightpaff}. This method leverages the per token KL divergence between the teacher- and student-model distributions to compute the training loss. To mitigate the tendency of student models to overemphasize low-probability regions in the teacher distribution, recent studies have proposed substituting the conventional forward KL divergence with reverse KL divergence \cite{gu2023minillm}. 

\paragraph{Learning without Forgetting} In traditional neural network frameworks, incrementally introducing new capabilities into multitask architectures typically requires access to all task datasets, which is often impractical due to the inaccessibility of historical data and the prohibitive computational costs associated with retraining \cite{caruana1997multitask}. In the context of convolutional neural network (CNN) classification tasks, a regularization strategy combining KL divergence with CE loss in a weighted formulation has been proposed to address catastrophic forgetting when the model capabilities are incrementally expanded\cite{li2017learning}. Empirical evaluations demonstrate that this approach achieves performance comparable to the upper bound established by joint training of all tasks simultaneously, offering a computationally efficient alternative to full retraining while mitigating the degradation of previously learned skills.

\begin{figure*}[h]
\centering
\includegraphics[width=1.0\textwidth, trim=10pt 0pt 0pt 0pt, clip]{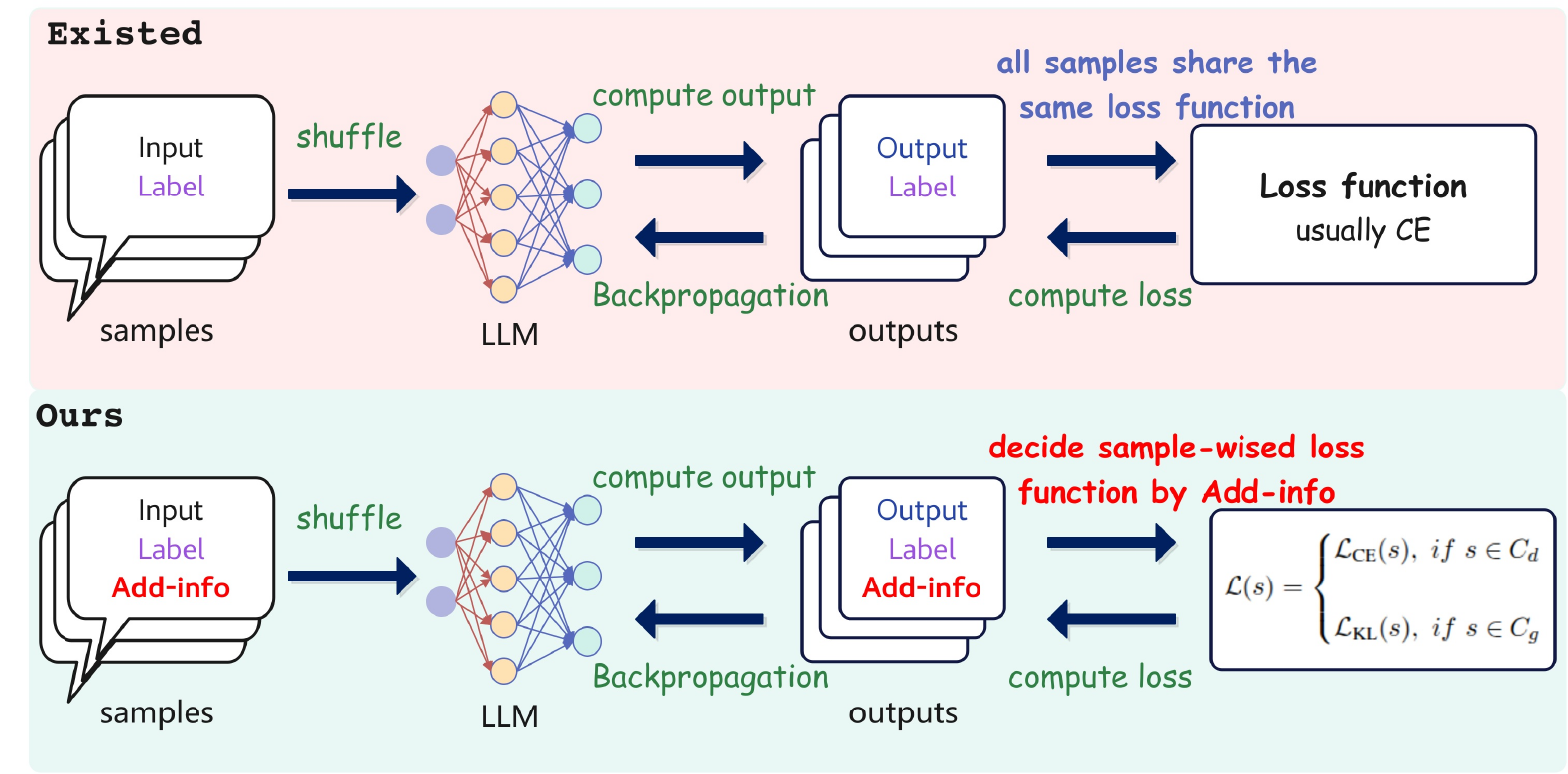} 
\caption{Schematic illustration of the MoL framework architecture. Unlike existed single-objective pre-training approaches, our MoL framework introduces an additional metadata input ("add-info") to distinguish between domain-specific and general corpora during training. This information determines the loss function selection: CE loss for domain corpora and KL divergence loss for general corpora (highlighted in red). The model's forward computation and backpropagation mechanisms retain the standard implementation pipeline of traditional LLMs.}
\label{figure0}
\end{figure*}

\begin{figure*}[h]
\centering
\includegraphics[width=1.08\textwidth, trim=20pt 5pt 5pt 40pt, clip]{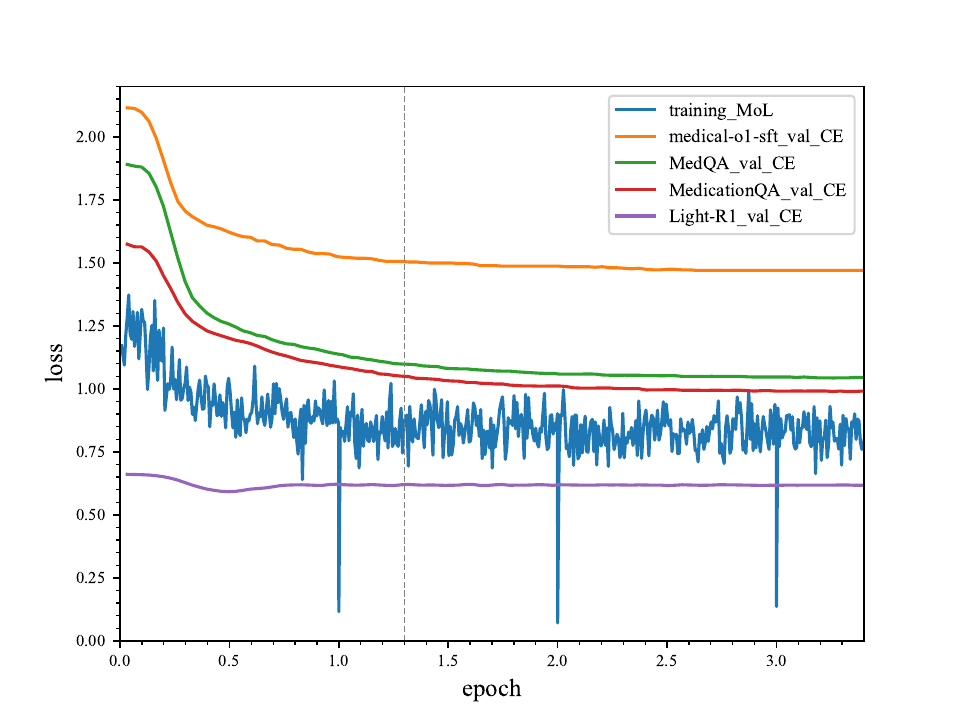} 
\caption{Training loss evolution across aggregated datasets and individual subsets, depicting CE loss dynamics for both domain-specific and general corpora. "train\_MoL" represents the loss on the training set under the MOL framework. The domain-corpus include medical-o1-sft, MedicationQA, and MedQA, while Light-R1 is the general-corpus. The "\_val" indicates the validation set, and "\_CE" denotes CE loss. The validation set's CE loss is calculated every 10 steps, resulting in a smoother curve compared to the training set loss. The CE loss for general-corpus remains nearly constant throughout training, while the domain-corpus exhibits a steady decline in loss until reaching convergence at nearly 1.3 epochs (marked by the dashed Line).}
\label{figure1}
\end{figure*}

\section{Methods}
The roles of domain-corpus and general-corpus in CPT fundamentally differ. Domain-corpus are primarily designed to enhance a model's domain-specific capabilities by fine-tuning its understanding and generation within specialized contexts. In contrast, general-corpus serve to preserve and refine the model's general capabilities, which are critical for both ensuring broad applicability in diverse tasks and maintaining foundational competencies such as chain-of-thought (CoT) reasoning \cite{jaech2024openai,xie2024preliminary}.

A notable method for preserving LLMs capabilities during training is the use of KL divergence as an objective function \cite{adler2021quantization}. Unlike traditional CE loss, which enforces deterministic "hard labels" by treating each token as an absolute target, KL divergence treats the output probability distribution of a base model as "soft labels" \cite{gu2023minillm}. This approach allows the target model to learn context-dependent generation patterns rather than memorizing fixed token sequences.

Thus our MoL framework integrates domain-specific and general corpora under a dual-perspective optimization strategy in CPT. For domain-corpus, we employ the CE loss with hard labels to enforce precise domain-specific knowledge acquisition. While for general-corpus, the KL divergence loss with soft labels is adopted to preserve the model's pre-existing generalization capabilities. The loss function $\mathcal{L}$ for each sequence $s$ in our MoL framework is formulated as follows:
\begin{align}
    \mathcal{L}(s)=\begin{cases}
        \mathcal{L}_{\text{CE}}(s),\,\,if \,\,s\in C_{d}, 
        \\
 \\ \mathcal{L}_{\text{KL}}(s),\,\,if \,\,s\in C_{g},
\end{cases}\label{MoL}
\end{align}
\begin{align}
\mathcal{L}_{\text{CE}}(s)=-\frac{1}{n_s}\sum_ilog\,p_{\theta}(s_i),\\
\mathcal{L}_{\text{KL}}(s)=\frac{1}{n_s}\sum_{i}\text{KL}[p_{\theta}||p_0](s_i),\label{KL}
\end{align}
where the average is performed over the total number of effective tokens ($n_s$). $p_0$ represents the probability distribution of the base LLM, and $p_{\theta}$ denotes the probability distribution of the CPT model parameterized by $\theta$. The sets $C_d$ and $C_g$ correspond to domain-specific and general corpora,  respectively. The specific operational workflow of our MoL framework is illustrated in Figure \ref{figure0}. This dual-loss architecture parallels human cognitive development: the KL divergence loss maintains alignment with foundational knowledge (like retaining language fundamentals during domain expertise acquisition), while the CE loss drives intentional knowledge expansion (similar to targeted skill development). The combination ensures that the model both preserves its general capabilities and systematically builds domain-specific expertise through complementary learning modes.

In practice, we introduce a small coefficient $\alpha$ to slightly adjust the final loss function to ensure training stability \cite{muller2019does}. Specifically, for the domain-corpus, the loss function is defined as 
\begin{align}
\mathcal{L} = (1-\alpha)\mathcal{L}_{\text{CE}} + {\alpha}\mathcal{L}_\text{KL},\label{domain}
\end{align}
while for the general-corpus, it is formulated as 
\begin{align}
\mathcal{L} = {\alpha}\mathcal{L}_{\text{CE}} + (1-\alpha)\mathcal{L}_\text{KL}.\label{general}
\end{align}
In our experiments, $\alpha$ is set to 0.01 unless otherwise specified.

In Equation \ref{KL}, we adopt the proposal of reverse KL divergence to mitigate overestimation of low-probability regions in the base model's output distribution \cite{gu2023minillm}. To further enhance regularization effectiveness, we introduce a cross-model probability aggregation scheme that jointly considers the probability distributions from both the base LLM and the CPT model for low-probability tokens, as formulated in Equation \ref{optimization_KL}. This optimization framework significantly reduces GPU memory consumption during KL divergence computation.

\begin{table*}
\centering
\begin{tabular}{clcc}
\toprule
\textbf{} & \textbf{} & \textbf{Qwen3-8B} & \textbf{+ D\&G 1:1} \\ 
\hline
\multirow{2}{*}{Domain} & MedQA & 74.87 & \textbf{77.25} \\ 
& MMLU-cli & 78.86 & \textbf{80.52} \\
\hline
\multirow{2}{*}{General} & C-Eval & 67.45 & \textbf{77.65} \\ 
& MMLU & 72.56 & \textbf{75.79} \\ 
\hline
\multirow{2}{*}{Coding} & MBPP & \textbf{68.40} & 68.00 \\ 
& HumanEval & \textbf{86.59} & 82.32 \\
\hline
\multirow{1}{*}{Math (Non-thinking)} & MATH-500 & \textbf{85.40} & 84.40 \\
\hline
\multirow{3}{*}{Math (Thinking)} & MATH-500 & 96.60 & \textbf{97.80} \\ 
& AIME 24 & 76.67 & \textbf{80.00} \\
& AIME 25 & 66.67 & \textbf{73.33} \\
\bottomrule
\end{tabular}
\caption{Performance comparison of various models across different task categories, including Domain, General, Coding, and Math tasks. The metrics represent the accuracy or performance scores achieved by each model on the respective tasks. The \textbf{D\&G 1:1} refers to the training of base model using an 
nearly equal mix of domain-specific and general corpora.}
\label{table1}
\end{table*}

\section{Experiments}
Our objective is to validate the efficacy of the MoL training framework through empirical evaluation in the medical domain. Specifically, we conduct CPT on a hybrid dataset comprising medical-domain corpora and open source corpus using an open source model architecture. Subsequently, we evaluate the trained model's performance in both the medical domain and general domain to assess the validity and robustness of the proposed framework under real-world application scenarios.

\paragraph{Base Model}
The open source Qwen3-8B model \cite{yang2025qwen3} serves as the base for CPT. Additionally, this model is utilized to compute KL divergence during training within the MoL framework for consistency in optimization.

\paragraph{Training}
Our training data are derived from two primary sources:\\
(1) Domain-corpus: Training sections from the medical-o1-sft \cite{chen2024huatuogpt}, MedicationQA \cite{abacha2019bridging}, and MedQA \cite{jin2021disease}.\\
(2) General-corpus: High-quality chain-of-thought data from the open source Light-r1 corpus \cite{wen2025light}, serving as a supplementary training resource for broader reasoning capabilities.

With applying chat-template and concatenating, we can adopt a mixed CPT strategy, enabling training both textual and QA samples within a single pipeline. \cite{qwen2.5} The total templated and concatenated domain-corpus comprises approximately 10,000 samples, with 100 samples randomly selected as the validation set. For the general-corpus, we use the stage1 part of the Light-r1 dataset with 76 K training samples, allowing for flexible adjustment of the ratio between domain-specific and general corpora during experimental design.

Experiments are conducted using the Low-rank adaptation (LoRA) training approach (with a rank of 64) \cite{hu2022lora}. All training is performed with the model's context length fixed at 8,192 tokens, ensuring compatibility with long input sequences. For LoRA training, we use a learning rate of 1e-4. All other hyperparameters remain consistent across experiments, including a cosine decay learning schedule with a warm-up ratio of 0.1 and a global batch size of 128.

\begin{table*}[ht!]
\centering
\begin{tabular}{clcccc}
\toprule
\textbf{} & \textbf{} & \textbf{D\&G 1:1} & \textbf{D\&G 1:0.5} & \textbf{D\&G 1:1.5} & \textbf{D\&G 1:2} \\ 
\hline
\multirow{2}{*}{Domain} & MedQA & \textbf{77.25} & 75.68 & 77.17 & 77.19 \\ 
& MMLU-cli & \textbf{80.52} & 79.22 & 79.87 & 79.59 \\
\hline
\multirow{2}{*}{General} & C-Eval & \textbf{77.65} & 77.04 & 77.20 & 77.59 \\ 
& MMLU & 75.79 & \textbf{76.48} & 74.73 & 69.36 \\ 
\hline
\multirow{2}{*}{Coding} & MBPP & 68.00 & 66.00 & 67.00 & \textbf{69.00} \\ 
& HumanEval & 82.32 & 81.10 & 80.49 & \textbf{83.54} \\
\hline
\multirow{1}{*}{Math (Non-thinking)} & MATH-500 & \textbf{84.40} & 84.20 & 81.20 & 80.60 \\
\hline
\multirow{3}{*}{Math (Thinking)} & MATH-500 & \textbf{97.80} & 96.40 & 97.20 & 96.40 \\ 
& AIME 24 & \textbf{80.00} & \textbf{80.00} & 76.67 & 70.00 \\
& AIME 25 & \textbf{73.33} & 70.00 & 70.00 & 66.67 \\
\bottomrule
\end{tabular}
\caption{Performance evaluation of model variants trained on Qwen3-8B across diverse task categories, including Domain, General, Coding, and Math tasks. The \textbf{D\&G} ratios indicate the adjusted proportions of domain-specific to general corpora used for training, showing the influence of these ratios on model performance across different tasks.}
\label{table2}
\end{table*}

\paragraph{Evaluation}We perform a comprehensive evaluation of the trained models. The evaluation focuses on its performance in terms of domain, general knowledge, mathematics, and coding capabilities. The evaluation dataset of the trained model contains these benchmarks:\\
\textbullet\textbf{ Domain Tasks:} We use benchmarks including predefined test set of MedQA \cite{jin2021disease}, including 3426 Chinese questions and 1273 US questions. And MMLU-cli, the medicine-related test data in MMLU \cite{hendrycks2020measuring}, including 134 Anatomy questions, 264 Clinical questions, 143 College biology questions, 172 college medicine questions, 99 Medical Genetic questions and 271 professional medicine questions.\\
\textbullet\textbf{ General Tasks:} MMLU \cite{hendrycks2020measuring} and C-Eval \cite{huang2023c} (5-shot) \\
\textbullet\textbf{ Coding tasks:}  MBPP \cite{austin2021program} and HumanEval \cite{chen2021evaluating}\\
\textbullet\textbf{ Math Tasks:} MATH-500 \cite{lightman2023let}, AIME 24 and AIME 25 \cite{aime}.

The evaluation of Domain Tasks, General Tasks, and Coding Tasks occurs under a non-thinking mode. In contrast, MATH-500 is assessed under both thinking and non-thinking modes. AIME 24 and AIME 25 are exclusively evaluated in thinking mode. For all models operating in thinking mode, we employ a sampling temperature of 0.6, a top-p value of 0.95, and a top-k value of 20. In the non-thinking mode for General, Coding, and Math Tasks, the sampling hyperparameters are configured as follows: temperature = 0.7, top-p = 0.8, top-k = 20, and presence penalty = 1.5. The settings of evaluation parameter above are fully consistent with the official Qwen3 \cite{yang2025qwen3}. For domain tasks evaluated in non-thinking mode, the sampling hyperparameters are set with a temperature of 0.01. For both thinking and non-thinking modes, the maximum output length is capped at 30,720 tokens. Non-thinking mode is achieved by setting "enable\_thinking=False".

\section{Results}
\subsection{Main Results}
\paragraph{Determination of Optimal Training Steps} We first conduct experiments on Qwen3-8B using a nearly 1:1 ratio of medical and general corpora to investigate the training dynamics of the MoL framework. As shown in Figure \ref{figure1}, the CE loss for general corpora remains nearly constant throughout training, due to the use of KL divergence as the loss function for these samples. In contrast, the CE loss for domain-specific corpora exhibits a consistent downward trend until convergence. This observation aligns precisely with our hypothesis that MoL can effectively enhance domain knowledge while preserving general language capabilities. Notably, we observe that all datasets' CE losses approach convergence at approximately 1.3 training epochs. This finding establishes a critical reference point for subsequent model comparisons, and therefore we standardize all evaluations at this epoch for fair performance assessment across different training paradigms.

\paragraph{Performance Evaluation at Convergence} The results of 1.3 training epochs are presented in Table \ref{table1}. Our model using the MoL training framework demonstrates superior performance over the base Qwen3-8B model across three critical dimensions: domain-specific capabilities, general abilities, and CoT reasoning. We also observe a significantly larger discrepancy in C-Eval performance scores before and after training, which was primarily attributed to insufficient instruction-following (IF) capability in the base model. This limitation leads to systematic misinterpretation of multiple-choice answers during evaluation. However, the implementation of the MoL training approach effectively resolves this issue, resulting in complete elimination of parsing errors in the CPT model. Detailed comparisons are presented in Appendix \ref{analysis_ceval}.

\begin{table*}[ht!]
\centering
\begin{tabular}{clccc}
\toprule
\textbf{} & \textbf{} & \textbf{D\&G 1:1} & \textbf{D\&G 1:1 CE} & \textbf{D\&G 1:1 ($\alpha=0.5$)} \\ 
\hline
\multirow{2}{*}{Domain} & MedQA & 77.25 & \textbf{77.57} & 73.19 \\ 
& MMLU-cli & \textbf{80.52} & 80.24 & 78.95 \\
\hline
\multirow{2}{*}{General} & C-Eval & \textbf{77.65} & 76.23 & 73.86 \\ 
& MMLU & 75.79 & 57.71 & \textbf{76.99} \\ 
\hline
\multirow{2}{*}{Coding} & MBPP & \textbf{68.00} & 62.20 & 64.80 \\ 
& HumanEval & 82.32 & 78.66 & \textbf{84.15} \\
\hline
\multirow{1}{*}{Math (Non-thinking)} & MATH-500 & \textbf{84.40} & 66.00 & 82.40 \\
\hline
\multirow{3}{*}{Math (Thinking)} & MATH-500 & \textbf{97.80} & 94.20 & 96.60 \\ 
& AIME 24 & \textbf{80.00} & 63.33 & 70.00 \\
& AIME 25 & \textbf{73.33} & 40.00 & 56.67 \\
\bottomrule
\end{tabular}
\caption{Performance evaluation of model variants trained on Qwen3-8B across diverse task categories, including Domain, General, Coding, and Math tasks. The \textbf{D\&G 1:1} corresponds to the definition provided in Table \ref{table1}. The \textbf{D\&G 1:1 CE} configuration utilizes CE as the loss function across all data. The final column represents results obtained with a $\alpha$ parameter of 0.5 in Equation \ref{domain} and \ref{general}.}
\label{table3}
\end{table*}%
Given that the open source medical corpus has probably already been exposed to Qwen3, further training on these domain-specific data may yield limited performance improvements in specialized medical tasks. To address this limitation and further validate the robustness of our MoL approach, we conducted additional experiments using a different foundational architecture and an internal corpus. The domain-to-general data ratio was carefully balanced by augmenting the general-domain component through threefold repetition of another internal general-corpus. This setup enabled us to maintain sufficient training scale while ensuring domain relevance. The results, as shown in Appendix \ref{more_results}, demonstrate that the proposed method achieves notable performance gains on the internal domain evaluation set compared to the baseline. Importantly, the model's general linguistic capabilities remain consistent with the base model's performance on standard benchmarks, confirming that domain adaptation does not come at the cost of foundational language proficiency.
\paragraph{Optimal Domain-to-General Corpus Ratio} Regarding the optimization problem of domain-to-general corpus ratio, we conduct extensive experiments using the medical domain-specific corpus as the fixed unit and vary the proportion of general-domain data (0.5, 1, 1.5, and 2) accordingly, as illustrated in Table \ref{table2}. The experimental results demonstrate that, for our MoL framework, a ratio near 1:1 achieves the most balanced performance between domain-specific and general language capabilities. This configuration consistently outperforms other tested ratios across most of the evaluation metrics, indicating that the optimal trade-off point for this specialized loss mechanism lies in maintaining approximately equal proportions of domain-specific and general corpora. 

\begin{figure*}[t]
\includegraphics[width=1\textwidth, trim=10pt 10pt 5pt 10pt, clip]{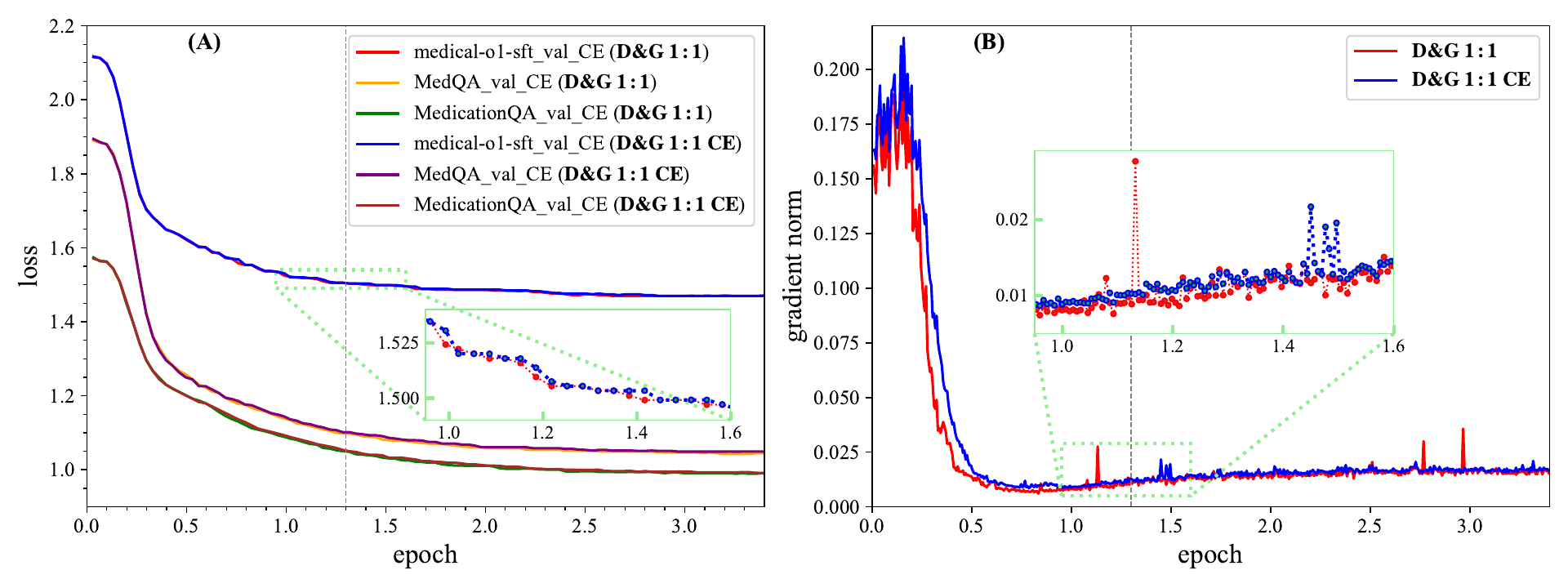} 
\caption{\textbf{(A)} Comparison of CE loss across various validation sets during training on Qwen3-8B. The main plot depicts the loss trends across different domain-corpus, including medical-o1-sft, MedQA and MedicationQA, while the inset offers a magnified view focusing on the performance of the medical-o1-sft under the two configurations. Specifically, \textbf{D\&G 1:1} refers to the setup described in Table \ref{table1}, while \textbf{D\&G 1:1 CE} denotes the alternative configuration that replaces KL divergence with CE throughout the training. The "\_val" indicates the validation set, and "\_CE" denotes CE loss. \textbf{(B)} Comparison of gradient norm during training. The main plot illustrates the gradient norm dynamics of the \textbf{D\&G 1:1} and \textbf{D\&G 1:1 CE} configurations across training epochs. While both configurations exhibit similar temporal evolution patterns, the \textbf{D\&G 1:1} consistently shows smaller magnitudes than \textbf{D\&G 1:1 CE} until convergence, after which the two curves align closely. The inset provides a zoomed-in view centered at 1.3 training epochs.}
\label{figure2}
\end{figure*}

\subsection{Ablation Studies}
\subsubsection{Influence of KL Divergence}
To evaluate the necessity of KL divergence in our framework, we conduct one control experiment under the optimal mixture ratio of 1:1 (Table \ref{table2}). This experiment replaces all KL divergence calculations with CE  counterparts. Our training results demonstrate significantly superior performance compared to this alternative, as detailed in Table \ref{table3}. The significant performance gains, particularly a 27.9\% increase in accuracy on the Math-500 benchmark in the non-think reasoning mode and an impressive 83.3\% improvement on the challenging AIME25 subset in the think mode, further highlight the effectiveness of integrating KL divergence into our training framework. These results not only reflect the superiority of our method, but also suggest that the alternative approach using CE may suffer from a degradation in generalization capability under the same experimental setup. This decline in general performance across different reasoning modes and benchmarks further supports the necessity of KL divergence in enhancing the robustness and adaptability of our model.

Regarding domain-specific capabilities, we observe that the CE losses of various domain-specific corpora under the MOL framework are closely aligned with those of the control experiment that employs CE exclusively, as shown in Figure \ref{figure2} \textbf{(A)}. This suggests that our method is on par with the traditional CE-based training approach in terms of domain adaptation and specialization. However, as previously discussed, our framework significantly outperforms the CE-only alternative in general reasoning tasks, particularly in complex and abstract reasoning scenarios.

Furthermore, we visualize the gradient magnitudes of both experiments during training, as presented in Figure \ref{figure2} \textbf{(B)}. While the two curves exhibit similar shapes, the gradients of the KL divergence-based framework remain consistently smaller than those of the CE-based alternative until convergence. Given that the control experiment replaces the KL divergence with CE, this observation implies that KL divergence contributes relatively less to the gradient on general corpora compared to CE, until training converges. Considering that the gradient from KL divergence is initially close to zero and gradually increases during training, this dynamic behavior suggests a form of a negative feedback regression mechanism \cite{zhao2018recommendations}. Consequently, the introduction of KL divergence within the MOL framework elegantly ensures the preservation of general reasoning capabilities, thereby demonstrating its necessity.

\subsubsection{Influence of Coefficient $\alpha$}
Furthermore, while the efficacy of KL divergence has been demonstrated, it is essential to justify whether the hyperparameter $\alpha$ in Equation  \ref{domain} and \ref{general} significantly influences the final outcomes. To investigate this, we configure $\alpha$ to 0.5, effectively assigning equal importance to domain-specific and general corpora. Our experiments reveal that a near-zero $\alpha$ value consistently yielded superior performance. The quantitative validation supporting this observation is detailed in Table \ref{table3}.

\section{Conclusion}
This study introduces the MoL framework, a dual-loss architecture that synergistically preserves general language capabilities while enhancing domain-specific performance through decoupled optimization. By applying CE loss for domain-corpus training and KL divergence for general-corpus alignment, the framework mitigates catastrophic forgetting in foundational skills while systematically integrating specialized knowledge. The 1:1 domain-to-general corpus ratio is empirically validated as optimal, demonstrating its ability to prevent overfitting while avoiding laborious and computationally intensive hyperparameter tuning processes. These contributions establish MoL as a principled, scalable solution for multi-domain language model training, offering both theoretical insights and practical deployment advantages in real-world heterogeneous scenarios.

\section*{Limitations}
While the MoL framework achieves notable improvements, one key limitation lies in the unexpected enhancement of AIME reasoning and IF capabilities through general KL divergence alignment on general-corpus. This phenomenon, though empirically observed, lacks a systematic analysis of its underlying mechanism. Further investigation is required to clarify its role in bridging general and specialized performance within the MoL framework.

\bibliography{sample}

\appendix
\onecolumn
\section{Optimization of KL Divergence}
\label{optimization_KL}
We introduce an additional regularization scheme by aggregating the probability values of low-probability tokens across both the base LLM and the CPT model. The aggregated probability distribution is then used for calculating the reverse KL divergence. This approach focuses the optimization process on aligning the high-probability regions of the base model's generation behavior, sharing conceptual parallels with physical systems analysis where attention is directed towards low-energy states that dominate system behavior \cite{schollwock2005density}.

For example, the vocabulary of Qwen3 exceeds 150,000 tokens \cite{yang2025qwen3}. However, we employ a dynamic probability truncation strategy to reduce the vast vocabulary size by focusing on the most probable tokens at each generation step. Specifically, given the base model's probability distribution, we retain the top $n$ most probable tokens (denoted as set $T$), preserving their individual probabilities. The remaining probabilities outside of $T$ are aggregated into a single residual probability mass, 
\begin{equation}
\begin{split}
p'_{0,\,residual}=\sum_{t\notin T} p_0(t),\\
p'_{\theta, residual}=\sum_{t\notin T} p_{\theta}(t),
\end{split}
\end{equation}
resulting in a reduced distribution $p'$ that maintains the total probability mass of 1. In our work, we set the parameter n to 31.

\section{Analysis of C-Eval Results}
\label{analysis_ceval}
We observe that the C-eval benchmark exhibits lower evaluation results for the Qwen3-8B model. Analysis of the evaluation results \ref{wrong} and \ref{right} (already translated into English) reveals that even in non-thinking mode with 5-shot examples, the model may sequentially analyze options, leading to incorrect identification of the first capital letter as the final answer selection. The trained model demonstrates improved IF capabilities, enabling more accurate output generation based on provided examples.

\section{More Results}
\label{more_results}
We evaluate the MoL framework using Qwen2.5-7B-Instruct as the base model, training on a 330M-token internal domain corpus and a 300M-token general corpus (100M-token Magpie-built corpus repeated three times). This configuration achieves a domain-to-general token ratio of approximately 1:1 while maintaining sufficient training scale. The general corpus was explicitly construct using the Magpie framework \cite{xu2024magpie}. Full-parameter training was implemented with a learning rate of 1e-5 and a batch size of 1024, while all other hyperparameters were aligned with those specified in the main text.

The results showed significant performance improvements on the internal domain evaluation set compared to the baseline, with no degradation in general linguistic capabilities. The model retained consistent performance on standard benchmarks, demonstrating the framework's ability to preserve foundational language skills during domain-specific training. Table \ref{table5} summarizes these findings.

\noindent 
\refstepcounter{algorithm}
\begin{tcolorbox}[
    colback=gray!10,
    colframe=black,
    coltitle=white,
    colbacktitle=black,
    fonttitle=\bfseries,
    title=Qwen3-8B Model Outputs on C-eval with 5-Shot Examples,
    boxrule=0.5mm,
    width=\textwidth,
    before=\medskip,
    after=\medskip,
]
\textbf{Wrong Result:} \\
\{\\
\hspace*{4ex}"prompt": [\\
\hspace*{8ex}...(previous shots)\\
\hspace*{8ex}\{\\
\hspace*{12ex}"role": "HUMAN",\\
\hspace*{12ex}"prompt": ...(omitted prompt),\\
\hspace*{8ex}\},\\
\hspace*{8ex}\{\\
\hspace*{12ex}"role": "BOT",\\
\hspace*{12ex}"prompt": "D",\\
\hspace*{8ex}\},\\
\hspace*{8ex}\{\\
\hspace*{12ex}"role": "HUMAN",\\
\hspace*{12ex}"prompt": "The following is a single-choice question from a logic exam in China...\textbackslash nAnswer: "\\
\hspace*{8ex}\}\\
\hspace*{4ex}],\\
\hspace*{4ex}"origin\_prediction": "This question tests an argument support type logical reasoning question. We need to find an option that can most effectively support the conclusion in the original text...\#\#\# Option analysis:\textbackslash n\textbackslash n\textcolor{red}{A}...",\\
\hspace*{4ex}"predictions": "\textcolor{red}{A}",\\
\hspace*{4ex}"references": "\textcolor{green}{C}"\\
\}
\label{wrong}
\end{tcolorbox}

\noindent 
\refstepcounter{algorithm}
\begin{tcolorbox}[
    colback=gray!10, 
    colframe=black, 
    coltitle=white, 
    colbacktitle=black, 
    fonttitle=\bfseries, 
    title=CPT Model Outputs on C-eval with 5-Shot Examples, 
    boxrule=0.5mm, 
    width=\textwidth, 
    before=\medskip, 
    after=\medskip, 
]
\textbf{Right Result:} \\
\{\\
\hspace*{4ex}"prompt": [\\
\hspace*{8ex}...(previous shots)\\
\hspace*{8ex}\{\\
\hspace*{12ex}"role": "HUMAN",\\
\hspace*{12ex}"prompt": ...(omitted prompt),\\
\hspace*{8ex}\},\\
\hspace*{8ex}\{\\
\hspace*{12ex}"role": "BOT",\\
\hspace*{12ex}"prompt": "D",\\
\hspace*{8ex}\},\\
\hspace*{8ex}\{\\
\hspace*{12ex}"role": "HUMAN",\\
\hspace*{12ex}"prompt": "The following is a single-choice question from a logic exam in China..."\textbackslash nAnswer: "\\
\hspace*{8ex}\}\\
\hspace*{4ex}],\\
\hspace*{4ex}"origin\_prediction": "\textcolor{green}{C}",\\
\hspace*{4ex}"predictions": "\textcolor{green}{C}",\\
\hspace*{4ex}"references": "\textcolor{green}{C}"\\
\}
\label{right}
\end{tcolorbox}


\begin{table*}
\centering
\begin{tabular}{clcc}
\toprule
\textbf{} & \textbf{} & \textbf{Qwen2.5-7B-Instruct} & \textbf{+ D\&G 1:1} \\ 
\hline
\multirow{5}{*}{Domain} & Concept-explanation & 58.33 & \textbf{67.96} \\
& Summarize & 39.84 & \textbf{53.50} \\ 
& Simple-QA & 48.67 & \textbf{57.70} \\
& Ops-FAQ & 17.37 & \textbf{63.59} \\
& Product-FAQ & 23.34 & \textbf{41.32} \\
\hline
\multirow{5}{*}{General} & C-Eval & \textbf{79.10} & 79.06 \\ 
& MMLU & 74.27 & \textbf{74.39} \\
& CMMLU & 78.67 & \textbf{78.68} \\
& BBH & \textbf{69.70} & 67.04 \\ 
& ‌HellaSwag & 81.87 & \textbf{81.91} \\ 
\hline
\multirow{2}{*}{Coding} & MBPP & \textbf{66.60} & 64.80 \\
& HumanEval & \textbf{81.10} & \textbf{81.10} \\
\hline
\multirow{2}{*}{Math} & MATH & \textbf{57.60} & 57.56 \\
& Gsm8k & 85.14 & \textbf{85.15} \\
\bottomrule
\end{tabular}
\caption{Performance comparison of various models across different task categories, including Domain, Business-related, Coding, General, and Math tasks. The \textbf{D\&G 1:1} corresponds to the definition provided in Table \ref{table1}.}
\label{table5}
\end{table*}

\end{document}